\documentclass[letterpaper, 10 pt, conference]{ieeeconf}
\IEEEoverridecommandlockouts

\usepackage[utf8]{inputenc}
\usepackage[english]{babel}
\usepackage{amsmath}
\usepackage{amsfonts}
\usepackage{amssymb}
\usepackage{bbm}
\usepackage{cite}

\usepackage{amsthm}

\usepackage{url}
\usepackage{mathtools}
\usepackage{comment}
\usepackage[keeplastbox]{flushend}

\usepackage{epsfig}
\usepackage{color}
\usepackage{tikz}
\usepackage{booktabs,tabularx}
\usepackage{multirow,arydshln}
\usepackage{algorithm}
\usepackage{algorithmic}
\usepackage{subcaption}
\usepackage{sidecap}

\newtheorem{definition}{Definition}[]
\newtheorem{problem}{Problem}[]

\newtheorem{assumption}{Assumption}

\newtheorem{theorem}{Theorem}[]

\newtheorem{lemma}{Lemma}[]

\newcommand\norm[1]{\left\lVert#1\right\rVert}
\newcommand{\argmax}{\arg\!\max}

\DeclareMathOperator{\E}{\mathbb{E}}

\def\R{{\mathbb{R}}}

\def\grset{{\cal X}}
\def\T{{\cal T}}
\def\srv{\mathbf{s}}
\def\omegarv{\boldsymbol{\omega}}
\def\ent{\mathcal{H}}
\def\info{\mathcal{I}}
\def\dkl{D_{\mathcal{KL}}}

\def\T{T}

\def\Next{\kern0.25ex\vcenter{\hbox{$\scriptstyle\bigcirc$}}\kern0.25ex}

\newcommand{\etal}{{\em et~al.~}}
\newcommand{\labeq}[1]{\stackrel{#1}{=}}
\newcommand{\labgeq}[1]{\stackrel{#1}{\ge}}
\newcommand{\lableq}[1]{\stackrel{#1}{\le}}

\newcommand{\RNum}[1]{\uppercase\expandafter{\romannumeral #1\relax}}

\newcommand{\probref}[1]{Problem~\ref{#1}}
\newcommand{\algref}[1]{Algorithm~\ref{#1}}
\newcommand{\asmpref}[1]{Assumption~\ref{#1}}

\newcommand{\sectref}[1]{Section~\ref{#1}}

\newcommand{\figref}[1]{Fig.~\ref{#1}}

\newcommand{\thmref}[1]{Theorem~\ref{#1}}

\title{\bf
Online Active Perception for Partially Observable \\ Markov Decision Processes with Limited Budget}
\author{Mahsa Ghasemi, Ufuk Topcu$^{}$ %
\thanks{Mahsa Ghasemi is with the Department of Electrical and Computer Engineering, University of Texas at Austin, Austin, TX 78712 USA. Ufuk Topcu is with the Department of Aerospace Engineering and Engineering Mechanics, University of Texas at Austin, Austin, TX 78712 USA.}%
\thanks{This work was supported in part by ONR grants N00014-19-1-2054 and N00014-18-1-2829, and DARPA grant D19AP00004.}%
}
\begin{document}

\maketitle

\begin{abstract}
	Active perception strategies enable an agent to selectively gather information in a way to improve its performance. In applications in which the agent does not have prior knowledge about the available information sources, it is crucial to synthesize active perception strategies at runtime. We consider a setting in which at runtime an agent is capable of gathering information under a limited budget. We pose the problem in the context of partially observable Markov decision processes. We propose a generalized greedy strategy that selects a subset of information sources with near-optimality guarantees on uncertainty reduction. Our theoretical analysis establishes that the proposed active perception strategy achieves near-optimal performance in terms of expected cumulative reward. We demonstrate the resulting strategies in simulations on a robotic navigation problem.
\end{abstract}
%%%%%%%%%%%%%%%%%%%%%%%%%%%%%%%%%%%%%%%%%%%%%%%%%%%%%%%%
%%%%%%%%%%%%%%%%%%%%%%%Introduction%%%%%%%%%%%%%%%%%%%%%
%%%%%%%%%%%%%%%%%%%%%%%%%%%%%%%%%%%%%%%%%%%%%%%%%%%%%%%%
\section{Introduction}\label{sec:intro}
An intelligent system should be able to exploit the available information in its surroundings toward better accomplishment of its task. However, in many applications in robotics and control, a decision-maker (called an agent) is not necessarily aware of the available information sources during a priori planning. For instance, consider an environment in which multiple agents, each with individual plans for their specific tasks, operate together. An agent may have no or only limited access to the behavioral model of other agents, and hence their observability of the environment and whether they are in the communication range. Nevertheless, at runtime, the agents may decide to exchange their information in order to enhance their performance.

In practical settings, the ability of an agent in gathering information is subject to budget constraints originating from power, communication, or computational limitations. If an agent decides to employ a sensor, it incurs a cost associated with the required power, or, if an agent decides to communicate with another agent, it incurs a communication cost. Such budget constraints accentuate the need for actively selecting a subset of available information that are most beneficial to the agent. We call this decision-making problem \textit{budget-constrained online active perception}.

We formulate budget-constrained online active perception for partially observable Markov decision processes (POMDPs). Computing an optimal policy for POMDPs that maximizes the expected cumulative reward, is generally PSPACE-complete~\cite{papadimitriou1987complexity}. This complexity result has led to design of numerous approximation algorithms. A well-known family of these approximate methods relies on point-based value iteration solvers~\cite{cheng1988algorithms,kurniawati2008sarsop,smith2012point}. Point-based solvers exploit the piecewise linearity and convexity~\cite{sondik1978optimal} of value function to approximate it as the maximum of a set of hyperplanes, each associated with a sampled belief point. It is provable that the error due to this approximation is bounded by a factor depending on the density of sampled belief points~\cite{pineau2006anytime}.

The combinatorial nature of selecting a subset of available information subject to budget constraints renders the task of finding an optimal solution NP-hard. We propose an efficient yet near-optimal online active perception strategy for POMDPs that aims to minimize the agent's uncertainty about the state while respecting the constraint. We prove the near-optimality of the proposed algorithm. Further, we evaluate the efficacy of the proposed solution for a robotic navigation task where the robot can communicate with unmanned aerial vehicles (UAVs) to better localize itself. 

%%%%%%%%%%%%%%%%%%%%%%%%%%%%%%%%%%%%%%%%
\subsection{Related Work}

Active perception has been studied in many applications including robotics~\cite{elfes1990occupancy,stone2006pixels,charrow2015active,best2016decentralised} and image processing~\cite{darrell1996active,vogel2007non}. A body of literature formalizes active perception as a reward-based task of a POMDP, enabling non-myopic decision-making. The reward-based treatment of perception has been employed for active classification~\cite{guo2003decision} and cooperative active perception~\cite{spaan2008cooperative,spaan2009decision,natarajan2015multi}.
Araya \etal~\cite{araya2010pomdp} introduce $\rho$POMDP model in which the reward is the entropy of the belief and Spaan \etal~\cite{spaan2015decision} propose POMDP-IR in which the reward depends on the accuracy of state prediction. In~\cite{satsangi2018exploiting}, the authors exploit the submodularity of value function for $\rho$POMDP and POMDP-IR to design a greedy maximization technique for finding a near-optimal active perception policy.
Our setting differs from the existing work in two aspects. First, we consider both planning and perception where the perception serves the planning objective. Second, we consider settings in which the perception model in only partially known in a priori planning.

An instance of active perception, considered in this paper, is that of dynamically selecting a subset of available information sources. The existing work on subset selection quantify usefulness of an information source by information-theoretic utility functions such as scalarizations of error covariance matrix of the estimated parameter~\cite{shamaiah2010greedy,hashemi2018randomized}, mutual information between the measurements and the parameter of interest, or entropy of the selected measurements~\cite{krause2007near,krause2014submodular}.
Given a specific utility function, selecting an optimal subset of information sources under constraint is a combinatorial problem~\cite{williamson2011design}.
However, if the utility function has properties such as monotonicity or (weak) submodularity, greedy algorithms can achieve near-optimal solutions with only polynomial number of function evaluations~\cite{nemhauser1978analysis,wang2016approximation,qian2017subset}. 
We use mutual information between the current state and the observations as the utility function. We obtain theoretical guarantee for the performance of the proposed generalized greedy maximization algorithm by exploiting monotonicity and submodularity of mutual information as well as linearity of cost constraint.

%%%%%%%%%%%%%%%%%%%%%%%%%%%%%%%%%%%%%%%%%%%%%%%%%%%%%%%%
%%%%%%%%%%%%%%%%%%%%%%%Notations&stuff%%%%%%%%%%%%%%%%%%
%%%%%%%%%%%%%%%%%%%%%%%%%%%%%%%%%%%%%%%%%%%%%%%%%%%%%%%%
\section{Preliminaries and Problem Statement}\label{sec:pre}
In this section, we provide an outline of the related concepts and definitions in order to formally state the problem.

%%%%%%%%%%%%%%%%%%%%%%%%%%%%%%%%%%%%%%%%
\subsection{Preliminaries}
We first overview the necessary background on partially observable Markov decision processes (POMDPs), point-based value iteration solvers, and properties of set functions.

%%%%%%%%%%%%%%%%%%%%%%%%%%%%%%%%%%%%%%%%
\subsubsection{POMDP}

A POMDP is a tuple $\mathcal{P}=(S,A,T,\Omega,O,R,\gamma)$, where $S$ is the finite set of states, $A$ is the finite set of actions, $T : S \times A \times S \rightarrow [0,1]$ is the probabilistic transition function, $\Omega$ is the set of observations, $O : S \times A \times \Omega \rightarrow [0,1]$ is the probabilistic observation function, and $\gamma \in [0,1]$ is the discount factor.
At each time step, the environment is in some state $s \in S$. The agent takes an action $a \in A$ that causes a transition to a state $s' \in S$ with probability $Pr(s'|s,a) = T(s,a,s')$. Then it receives an observation $\omega \in \Omega$ with probability $Pr(\omega|s',a) = O(s',a,\omega)$, and a scalar reward $R(s,a)$. 

The belief of the agent at each time step, denoted by $b_t$ is the posterior probability distribution of states given the history of previous actions and observations, i.e., ${h_t = (a_0, \omega_1, a_1, \ldots, a_{t-1}, \omega_t)}$. 
A well-known fact is that due to Markovian property, a sufficient statistics to represent history of actions and observations is the belief \cite{aastrom1965optimal,smallwood1973optimal}. Given the initial belief $b_0$, the following update equation holds between previous belief $b$ and the belief $b_b^{' a,\omega}$ after taking action $a$ and receiving observation $\omega$:
\begin{equation}\label{eq:up1}
b_b^{' a,\omega}(s') = \frac{O(s',a,\omega) \sum_{s} T(s,a,s') b(s)}
{\sum_{s'} O(s',a,\omega) \sum_{s} T(s,a,s') b(s)}.
\end{equation}

The agent's objective is to find a pure policy that maximizes its expected discounted cumulative reward denoted by $\E[\sum_{t=0}^{\infty} \gamma^t R(s_t,a_t) | b_0]$. A pure policy is a mapping from belief to actions $\pi : B \rightarrow A$, where $B$ is the set of belief states. Note that $B$ constructs a $(|S|-1)$-dimensional probability simplex which we indicate by $\Delta_B$.

%%%%%%%%%%%%%%%%%%%%%%%%%%%%%%%%%%%%%%%%
\subsubsection{Point-Based Value Iteration}

POMDP solvers apply value iteration~\cite{sondik1978optimal}, a dynamic programming technique, to find the optimal policy. Let $V$ be a value function that maps beliefs to values in $\R$ that represent the expected discounted cumulative reward for a given belief. The following recursive expression holds for $V$:
\begin{equation}\label{eq:backup}
\begin{aligned}
V_t(b) = \max_{a} \Bigg(& \sum_{s \in S} b(s)R(s, a) +\\
&\gamma \sum_{\omega \in \Omega} Pr(\omega|b,a) V_{t-1}(b_b^{' a,\omega})\Bigg).
\end{aligned}
\end{equation}
The value iteration process converges to the optimal value function which satisfies the Bellman's optimality equation~\cite{bellman1957markovian}. Then, an optimal policy can be derived from the optimal value function.
An important outcome of~\eqref{eq:backup} is that at any horizon, the value function is piecewise linear and convex~\cite{smallwood1973optimal} and hence, can be represented by a finite set of hyperplanes. Each hyperplane is associated with an action. Let $\alpha$'s to denote the corresponding vectors of the hyperplane parameters and let $\Gamma_t$ to be the set of $\alpha$ vectors at horizon $t$. Then,
\begin{equation}\label{eq:alpha}
V_t(b) = \max_{\alpha \in \Gamma_t} \alpha \cdot b,
\end{equation}
where $\cdot$ indicates the dot product of the two vectors. Additionally, the action corresponding to the optimal $\alpha$ in~\eqref{eq:alpha} determines the optimal action at $b$.
This representation of the value function has motivated approximate point based solvers to try to approximate the value function by updating the hyperplanes over a finite set of sampled belief points.

Generic point-based solvers consist of three main steps, namely sampling, backup, and pruning. These steps are applied repeatedly until a desired convergence criterion for the value function is realized. For the sampling step, different approaches exist including discretization of the belief simplex and adaptive sampling techniques~\cite{pineau2006anytime,kurniawati2008sarsop,smith2012point}.
The backup step follows the standard Bellman backup operation. More specifically, one can rewrite~\eqref{eq:backup} using~\eqref{eq:alpha} to obtain:
\begin{equation*}
\begin{aligned}
V_t(b) = &\max_{a} \Bigg( \sum_{s \in S} b(s)R(s, a) + 
\sum_{\omega \in \Omega}
\max_{\alpha \in \Gamma_{t-1}} \\ 
& \sum_{s \in S} \sum_{s' \in S} \alpha(s') O(s',a,\omega) T(s,a,s') b(s) \Bigg),
\end{aligned}
\end{equation*}
where $\Gamma_{t-1}$ is the set of $\alpha$ vectors from previous iteration. Let $B_t$ to denote the current set of sampled belief points. The Bellman backup operator on $B_t$ is performed through the following procedure \cite{pineau2006anytime}:
\begin{equation*}
\begin{aligned}
\text{Step 1:} \;&
\text{For all } a \in A: \quad
\Gamma_t^{a,*} \leftarrow 
\alpha^{a,*}(s) = R(s,a) \\
\text{Step 2:} \;&
\text{For all } a \in A, \alpha \in \Gamma_{t-1}, \text{ and } \omega \in \Omega: \quad \\
&\Gamma_t^{a,\omega} \leftarrow 
\alpha^{a,\omega}(s) = \gamma \sum_{s' \in S} O(s',a,\omega) T(s,a,s') \alpha(s') \\
\text{Step 3:} \;&
\text{For all } a \in A, \text{ and } b \in B_{t}: \quad \\
&\Gamma_t^{b,a} \leftarrow \alpha^{b,a} = \alpha^{a,*} + \sum_{\omega \in \Omega} \argmax_{\alpha \in \Gamma_t^{a,\omega}} \alpha \cdot b \\
\text{Step 4:} \;&
\text{For all } b \in B_{t}: \quad
\alpha^b = \argmax_{\alpha \in \Gamma_t^{b,a}, a \in A} \alpha \cdot b \\
\text{Step 5:} \;&
\Gamma_t = \bigcup_{b \in B_t} \alpha^b
\end{aligned}
\end{equation*}
where $\Gamma_{t}$ is the new set of $\alpha$ vectors. Lastly, in the pruning step, the $\alpha$ vectors that are dominated by other $\alpha$ vectors are removed to simplify next round of computation~\cite{araya2010pomdp}. 

%%%%%%%%%%%%%%%%%%%%%%%%%%%%%%%%%%%%%%%%
\subsubsection{Properties of Set Functions}

Since the proposed active perception algorithm is founded upon the theoretical results from the field of submodular optimization for set functions, here, we overview the necessary definitions.
Let $\grset$ to denote a ground set and $f$ a set function that maps an input set to a real number.
\begin{definition}
	\label{def:mon}
	A set function $f : 2^\grset \rightarrow \mathbb{R}$  is monotone nondecreasing if $f(\T_1)\leq f(\T_2)$ for all $\T_1 \subseteq \T_2 \subseteq \grset$.
\end{definition}

\begin{definition}
	\label{def:submod}
	A set function $f : 2^\grset \rightarrow \mathbb{R}$ is submodular if 
	\begin{equation*}
	f(\T_1 \cup \{i\})-f(\T_1) \geq f(\T_2 \cup \{i\})-f(\T_2)
	\end{equation*}
	for all subsets $\T_1 \subseteq \T_2 \subset \grset$ and $i \in \grset \backslash \T_2$. The term $f_i(\T_1)=f(\T_1 \cup \{i\})-f(\T_1)$ is the marginal value of adding element $i$ to set $\T_1$.
\end{definition}
Monotonicity states that adding elements to a set increases the function value while submodularity refers to diminishing returns property.

%%%%%%%%%%%%%%%%%%%%%%%%%%%%%%%%%%%%%%%%
\subsection{Problem Statement}

In this paper, we consider an agent whose interaction with the environment, i.e., stochastic transitions and observations, is captured by a POMDP. In addition to a priori known observations captured by the POMDP, during runtime, the agent can further collect auxiliary observations, e.g., by means of communicating with other nearby agents. However, there is a budget constraint, such as limited communication bandwidth or limited communication power, on the auxiliary information gathering. Therefore, the agent must pick (or activate) a subset of auxiliary information sources that maximally increase its expected reward in the future while respecting the constraint. We formally state the problem next.

\begin{problem}\label{pr:1}
	Consider a POMDP $\mathcal{P}=(S,A,T,\Omega,O,R,\gamma)$ with initial belief $b_0$. Let set ${\Omega_t^{aux} = \Omega^1 \times \Omega^2 \times \ldots \times \Omega^{n_t}}$ to denote $n_t$ auxiliary observations available at time step $t$, with associated costs of ${c^1_t,c^2_t,\ldots,c^{n_t}_t}$, and an upper bound $\bar{c}_t$ on the cost. Also, let ${I_t = \{\iota = (i_1, i_2, \ldots, i_k) | i_j,k \in \{1,2,\ldots,n_t\} \}}$ to represent the power set obtained from $\Omega_t^{aux}$.
	In a priori planning, we aim to compute a pure belief-based policy $\pi: B \to A$ that maximizes the expected discounted cumulative reward, i.e,
	\begin{equation*}
		\pi^* = \argmax_{\pi} \: \E[\sum_{t=0}^{\infty} \gamma^t R(s_t,\pi(b_t)) | b_0].
	\end{equation*}
	Furthermore, at runtime, we aim to compute an active perception policy
	$\mu_t: B \to I_t$ that given current belief $b_t$, maximizes the expected discounted cumulative reward in the future while respecting the cost constraint, i.e.,
	\begin{equation*}
	\begin{aligned}
	\mu_t^* &= \argmax_{\mu_t} \: \E[\sum_{t}^{\infty} \gamma^t R(s_t,\pi(b_t)) | b_t] \\
	&\text{such that } \:
	\sum_{\stackrel{i \in \iota}{\iota = (i_1, i_2, \ldots, i_k) \in I_t}} c^{i}_t \le \bar{c}_t.
	\end{aligned}
	\end{equation*}
\end{problem}

%%%%%%%%%%%%%%%%%%%%%%%%%%%%%%%%%%%%%%%%%%%%%%%%%%%%%%%%
%%%%%%%%%%%%%%%%%%%%%%%%%%%Algorithm%%%%%%%%%%%%%%%%%%%%
%%%%%%%%%%%%%%%%%%%%%%%%%%%%%%%%%%%%%%%%%%%%%%%%%%%%%%%%
\section{Online Active Perception with\\ Limited Budget}\label{sec:alg}
\probref{pr:1} consists of two stages. 
The first stage is an a priori planning based on the POMDP model. We resort to point-based value iteration (see \sectref{sec:pre}) to compute a near-optimal policy $\hat{\pi}$ for this planning problem. As discussed earlier, various heuristics for adaptive sampling of belief points have been developed. The core idea of these methods is to guide the sampling toward the reachable subspace of the belief simplex $\Delta_B$. Nevertheless, since the reachable belief points depend on possible observations and the agent is not aware of auxiliary observations a priori, we propose a uniform sampling of the belief simplex. While uniform sampling is not as efficient as that of adaptive sampling for large POMDPs, it ensures coverage of the whole belief space.
The second stage of the problem is an online computation of an optimal subset of information sources with respect to expected future reward while complying with the cost constraint. To that end, we design a generalized greedy strategy, to be applied at each time step, which is computationally efficient and achieves near-optimal guarantees. 
Before introducing the algorithm, we state the following assumption regarding dependency of observations from the auxiliary information sources. 
\begin{assumption}\label{asmp:ind}
	We assume that the observations from the information sources are mutually independent given the current state and the previous action, i.e., 
	\begin{equation*}
	\begin{aligned}
		&\forall I, J \subseteq \{1,2,\ldots,n\}, I \cap J = \emptyset: \\ 
		&Pr(\bigcup_{i \in I} \omega^{i},\bigcup_{j \in J} \omega^{j}|\srv,a) = Pr(\bigcup_{i \in I} \omega^{i}|\srv,a)Pr(\bigcup_{j \in J} \omega^{j}|\srv,a).
	\end{aligned}
	\end{equation*}
\end{assumption}

Let $b_b^{' a,\omega}(s')$ to denote the updated belief after taking action $a$ and receiving observation $\omega$. Assume the agent then picks a perception action corresponding to ${\iota = (i_1, i_2, \ldots, i_k)}$ and receives an auxiliary observation ${\bar{\omega} = (\omega^{i_1},\omega^{i_2},\ldots,\omega^{i_k},)}$. Then, if \asmpref{asmp:ind} holds, according to Bayes' theorem, the agent's belief will be further updated by the following rule:
\begin{equation}\label{eq:up2}
	b_{b'}^{'' a,\iota,\bar{\omega}}(s'') =
	\frac{\prod_{i \in \iota} O_i(s'',a,\omega^i)b'(s'')}
	{\sum_{s''} \prod_{i \in \iota} O_i(s'',a,\omega^i)b'(s'')},
\end{equation}
where $O_i(s'',a,\omega^i) = Pr(\omega^i|s'',a,\iota)$.

%%%%%%%%%%%%%%%%%%%%%%%%%%%%%%%%%%%%%%%%
\subsection{Proposed Generalized Greedy Algorithm}

To quantify utility of information sources, we use mutual information between the state and auxiliary informations. Mutual information between two random variables is a positive and symmetric measure of their dependence and is defined as:
\begin{equation*}
\info(\boldsymbol{x};\boldsymbol{y}) = 
\sum_{x,y} p_{\boldsymbol{x},\boldsymbol{y}}(x,y) \log \frac{p_{\boldsymbol{x},\boldsymbol{y}}(x,y)}{p_{\boldsymbol{x}}(x) p_{\boldsymbol{y}}(y)}.
\end{equation*}
Mutual information, due to its monotonicity and submodular characteristics, has inspired many subset selection algorithms~\cite{krause2014submodular}. The mutual information between the state and the auxiliary informations is closely related to the change in the entropy of the state after receiving the additional observations, as expressed by the following equation:
\begin{equation}\label{eq:inf-ent}
	\info(\srv;\bigcup_{i \in \iota} \omegarv^i) = \ent(\srv)-\ent(\srv|\bigcup_{i \in \iota} \omegarv^i).
\end{equation}
For a discrete random variable $\boldsymbol{x}$, the entropy is defined as $\ent(\boldsymbol{x}) = -\sum_{i} p(x_i) \log p(x_i)$ and captures the amount of uncertainty. Therefore, intuitively, maximizing the mutual information is equivalent to minimizing the uncertainty in the state. Minimizing the state uncertainty is the goal of perception actions as it leads to higher expected reward in the future. Notice that the entropy is strictly concave on $\Delta_B$~\cite{cover2012elements}. Hence, minimizing the entropy pushes the belief toward the boundary of the simplex that due to convexity of the value function, possesses higher value. That being the case, in order to select the optimal perception action, we define the objective function as the following set function:
\begin{equation}\label{eq:gr-f}
	f(\iota) = \info(\srv;\bigcup_{i \in \iota} \omegarv^i) 
	= \ent(\srv) - \ent(\srv|\bigcup_{i \in \iota} \omegarv^i),
\end{equation}
and aim to compute $\iota^*$ by solving the following discrete optimization problem:
\begin{equation}\label{eq:gr-o}
	\iota^* = \argmax_{\iota} f(\iota) \quad
	\text{such that } 
	\sum_{\stackrel{i \in \iota}{\iota = (i_1, i_2, \ldots, i_k) \in I_t}} c^{i}_t \le \bar{c}_t.
\end{equation}

Note that $\ent(\srv)$ is constant and does not affect the selection procedure. Furthermore, $\ent(\srv|\bigcup_{i \in \iota} \omegarv^i)$ yields the expected value of entropy over all possible realizations of observations and can be computed via:
\begin{equation}
\begin{aligned}
\ent&(\srv | \bigcup_{i \in \iota} \omegarv^i) = -
\sum_{\omega^{i_1} \in \Omega^{i_1}} \ldots \sum_{\omega^{i_k} \in \Omega^{i_k}} \sum_{s \in S} \Bigg(b(s) \prod_{i_j \in \iota}\\ 
&O_{i_j}(s,a,\omega^{i_j}) 
\log \left(\frac{b(s) \prod_{i_j \in \iota} O_{i_j}(s,a,\omega^{i_j})}
{\sum_{s' \in S} b(s') \prod_{i_j \in \iota} O_{i_j}(s',a,\omega^{i_j})}\right) \Bigg).
\end{aligned}	
\end{equation}

At each time step, there is $2^n$ possible perception actions $\iota$ with their associated costs. Finding an optimal subset of information sources with respect to \eqref{eq:gr-o} is a combinatorial optimization problem and is NP-hard~\cite{williamson2011design}. Hence, we propose an approximate solution based on greedy maximization schemes.
The proposed greedy algorithm, outlined in \algref{alg:gr}, is founded upon the idea of generalized greedy algorithm in~\cite{lin2010multi}. 
\begin{algorithm}[h]
	\caption{Perception policy as a generalized greedy scheme}
	\begin{algorithmic}[1]
		\STATE \textbf{Input:} POMDP $\mathcal{P}=(S,A,T,\Omega,O,R,\gamma)$, Current belief $b$, Action $a$, Auxiliary information $\Omega_t^{aux}$ with costs ${c^1_t,c^2_t,\ldots,c^{n_t}_t}$, Cost constraint $\bar{c}_t$, Scaling factor $\beta>0$.
		\STATE \textbf{Output:} Perception action $\iota_t^g$.
		\STATE Initialize $\grset = \{1,2,\ldots,n_t\}$, $\tilde{\grset} = \grset$, $\tilde{\iota} = \emptyset$.
		\WHILE {$\tilde{\grset} \neq \emptyset$}
			\STATE $\displaystyle j^* = \argmax_{j \in \tilde{\grset} \backslash \tilde{\iota}} 
			\frac{\ent(\srv|\bigcup_{i \in \tilde{\iota}} \omegarv^i) -\ent(\srv|\bigcup_{i \in \tilde{\iota} \cup \{j\}} \omegarv^i)}
			{{(c_t^{j})}^\beta}$
			\IF {$\displaystyle \sum_{i \in \tilde{\iota}} c^{i}_t + c^{j^*}_t \le \bar{c}_t$}
				\STATE $\tilde{\iota} \leftarrow \tilde{\iota} \cup \{j^*\}$
			\ENDIF
			\STATE $\tilde{\grset} \leftarrow \tilde{\grset} \backslash \{j^*\}$
		\ENDWHILE
		\STATE $\displaystyle j^*_1 = \argmax_{j \in \grset, c^{j}_t \le \bar{c}_t} -\ent(\srv|\omegarv^j)$
		\STATE $\displaystyle \iota_t^g = \argmax_{\iota \in \{\tilde{\iota},\{j^*_1\}\}} -\ent(\srv|\bigcup_{i \in \iota} \omegarv^i)$
		\RETURN $\iota_t^g$.
	\end{algorithmic}
	\label{alg:gr}
\end{algorithm}
\captionsetup[figure]{name={Fig.}}
\begin{SCfigure}[][b]
	\centering
	\includegraphics[width=0.23\textwidth]{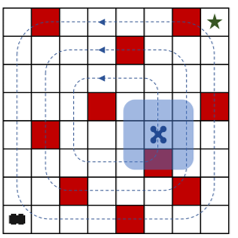}
	\caption{The robot aims to reach the target state (starred) while avoiding the obstacles (dark cells) in the map. UAVs periodically patrol the dashed paths and can view their nearby area (shaded area). The robot can ask for information from UAVs to better localize itself.}
	\label{fig:sim}
\end{SCfigure}
\begin{figure*}[t]
	\centering
	\begin{subfigure}[t]{0.18\textwidth}
		\includegraphics[width=\textwidth]{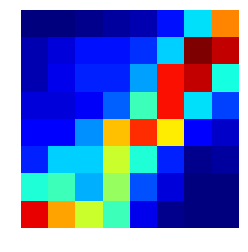}
		\caption{No auxiliary observations}
	\end{subfigure}\hfill
	\begin{subfigure}[t]{0.18\textwidth}
		\includegraphics[width=\textwidth]{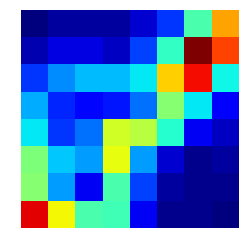}
		\caption{Random selection of 2 observations}
	\end{subfigure}\hfill
	\begin{subfigure}[t]{0.18\textwidth}
		\includegraphics[width=\textwidth]{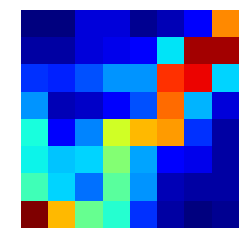}
		\caption{Greedy selection of 2 observations}
	\end{subfigure}\hfill
	\begin{subfigure}[t]{0.088\textwidth}
		\includegraphics[width=\textwidth]{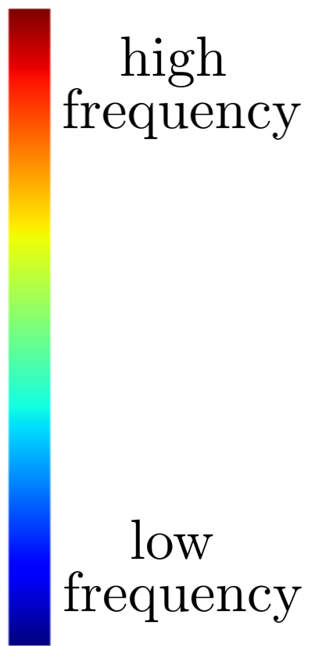}
		\caption*{}
	\end{subfigure}
	\caption{ The frequency of visiting states when using different online active perception methods.}\label{fig:heatmap}
\end{figure*}
The algorithm takes as input the agent's belief and action along the current set of available auxiliary informations. Then it iteratively adds elements from the ground set (set of all information sources) whose marginal gain with respect to $f$, scaled by the added cost, is maximal and terminates when no more elements can be added due to the constraint. Parameter $\beta$ is a scaling factor of the cost which can be adjusted to calibrate the effect of cost for a particular problem. The output set is the superior of the constructed subset and the best singleton subset. 

%%%%%%%%%%%%%%%%%%%%%%%%%%%%%%%%%%%%%%%%
\subsection{Theoretical Analysis}

Next, we theoretically analyze the performance of the proposed online active perception algorithm. The following lemma states the required properties of the objective function to prove near-optimality result.
\begin{lemma}
	Let $\Omega = \{\omegarv^1,\omegarv^2,\ldots,\omegarv^n\}$ to represent a set of observations of the state $\srv$ for which \asmpref{asmp:ind} holds. Then, $f(\iota)$, defined in \eqref{eq:gr-f}, realizes the following properties: 
	\begin{enumerate}
		\item $f(\emptyset) = 0$,
		\item $f$ is monotone nondecreasing, and
		\item $f$ is submodular.
	\end{enumerate}
\end{lemma}
The proof of the lemma follows from submodularity of conditional entropy \cite{ko1995exact} and its monotonicity.
The above lemma enables us to establish the approximation factor using the analysis in~\cite{lin2010multi}.
\begin{theorem}\label{thm:gr-app}
	Let $\iota^*$ to denote the optimal subset of observations obtained from the optimization problem in~\eqref{eq:gr-o}, and $\iota^g$ to denote the output of \algref{alg:gr} for $\beta=1$. Then, the following performance guarantee holds:
	\begin{equation}
		\info(\srv;\bigcup_{i \in \iota^g} \omegarv^i) \ge 
		\left(1-\frac{1}{\sqrt{e}}\right) \info(\srv;\bigcup_{i \in \iota^*} \omegarv^i).
	\end{equation}
\end{theorem}

\thmref{thm:gr-app} proves that the mutual information obtained by the generalized greedy algorithm is close to that of optimal solution in \eqref{eq:gr-o}. Nevertheless, we need to analyze the near-optimality of the proposed online active perception policy compared to $\mu_t^*$ in \probref{pr:1}. 
To that end, we show that the expected distance between the two belief points from greedy and optimal perception actions is bounded. 
Using this fact, we prove that the value loss is bounded as well.

\begin{theorem}\label{thm:dist}
	Let $b$ to denote the agent's current belief and $a$ to denote its last action.
	Further, let $\iota^g$ and $\iota^*$ to be the greedy perception action and the optimal action, respectively. Then, it holds that
	\begin{equation*}
		\E [\norm{b^g - b^*}_1] \le 
		\sqrt{\frac{2}{\sqrt{e}}\E_{\bigcup_{i \in \iota^*} \omegarv^i}\left[\dkl(p^* \| p^0)\right]},
	\end{equation*}
	where $b^*$ and $b^g$ are the updated beliefs according to \eqref{eq:up2}.
\end{theorem}

Now, we can use \thmref{thm:dist} to bound the value loss in the objective function in \probref{pr:1}.
\begin{theorem}\label{thm:val}
	Instate the notation and hypothesis of \thmref{thm:dist}. Additionally, let $V$ to be the computed value function for POMDP. It holds that:
	\begin{equation*}
		\E [V(b^g) - V(b^*)] \le \delta \frac{\max\left\{|R_{max}|,|R_{min}|\right\}}{1-\gamma},
	\end{equation*}
	where $\delta$ is the right hand side of the inequality in \thmref{thm:dist}.
\end{theorem}

%%%%%%%%%%%%%%%%%%%%%%%%%%%%%%%%%%%%%%%%%%%%%%%%%%%%%%%%
%%%%%%%%%%%%%%%%%%%%%%%Simulations%%%%%%%%%%%%%%%%%%%%%%
%%%%%%%%%%%%%%%%%%%%%%%%%%%%%%%%%%%%%%%%%%%%%%%%%%%%%%%%
\section{Simulation Results}\label{sec:sim}
We evaluate the proposed online active perception algorithm in a robotic navigation task. To that end, we implement a simple point-based value iteration solver that uses a fixed set of belief points. The belief points are uniformly distributed over $\Delta_B$ and their associated $\alpha$ vectors are initialized by $\frac{1}{1-\gamma} \text{min}_{s,a} R(s,a) \mathbf{1}_{|S|}$~\cite{shani2013survey}. We run the solver until the \mbox{$\ell_1$-norm} distance between value functions in two consecutive iterations falls below a predefined threshold of 0.001 or a maximum iteration number of 1000 is reached.
We implement the proposed generalized greedy selection algorithm as well as a random selection algorithm that selects a subset of information sources, uniformly at random.
After learning the policy from the solver, we apply the online active perception policies for 50 Monte Carlo simulation runs.\footnote{The code is available at https://github.com/MahsaGhasemi/greedy-perception-POMDP} 

The robotic navigation scenario models a robot in a $8 \times 8$ grid map whose objective is to reach a goal state while avoiding the obstacles in the environment, see \figref{fig:sim}. The goal state has a reward of 10, obstacle cells have a reward of -5, and other cells have a reward of -1. The navigation actions of the robot are $A = \{up,right,down,left,stop\}$. The robot's transitions are probabilistic due to possible actuation errors with 0.7 probability of taking the correct action.
The robot has an inaccurate sensor as well that can localize it correctly with probability 0.5. In addition to the robot, there are 12 UAVs that are patrolling the area in periodic motions. The field of view of each UAV is a $3 \times 3$ area. 
At each time step, the robot can select some of the UAVs and ask them to send their information regarding the state of the robot. However, note that the observation model of UAVs is time-varying and changes based on their location. Besides, the robot does not know the policies of UAVs during planning time. We assume that the cost of communicating with each UAV is the same. At each time step, the cost constraint allows communication with at most 2 UAVs.

\begin{figure}[t]
	\centering
	\includegraphics[width=0.37\textwidth]{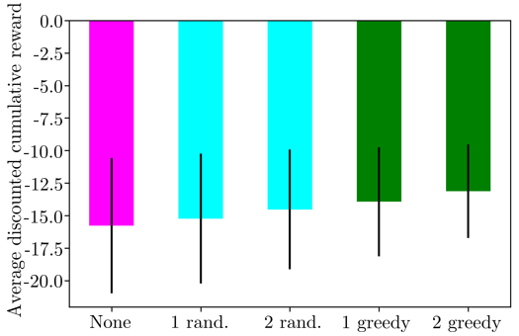}
	\caption{ The average discounted cumulative reward over 50 runs for each perception policy. The solid lines depict the corresponding standard deviations.}
	\label{fig:rew}
	\vspace{-0.5cm}
\end{figure}

We first find a planning policy via the implemented point-based solver. Next, we let the robot to run for a horizon of 40 steps, with no auxiliary information, with random selection of information sources, and with the proposed generalized greedy selection based on mutual information. We terminate the simulations once the robot reaches the goal. \figref{fig:heatmap} illustrates the normalized frequency of visiting each state for each perception algorithm. No use of auxiliary informations leads to worst performance as it visits the obstacle cells frequently. Random addition of auxiliary information sources improves the performance since it results in better obstacle avoidance. However, the best obstacle avoidance performance is for the proposed generalized greedy algorithm and it shows more concentration around the optimal path.
\figref{fig:rew} demonstrates the discounted cumulative reward, averaged over 50 Monte Carlo runs, for all three policies, i.e., no auxiliary information, random selection of 1 and 2 information sources, and greedy selection of 1 and 2 information sources. It can be seen that the generalized greedy selection scheme obtains the highest reward.

%%%%%%%%%%%%%%%%%%%%%%%%%%%%%%%%%%%%%%%%%%%%%%%%%%%%%%
%%%%%%%%%%%%%%%%%%%%%Conclusion%%%%%%%%%%%%%%%%%%%%%%%
%%%%%%%%%%%%%%%%%%%%%%%%%%%%%%%%%%%%%%%%%%%%%%%%%%%%%%
\section{Conclusion} \label{sec:concl}
We studied online active perception for POMDPs where at each time step, the agent can pick a subset of available information sources, under a budget constraint, to enhance its belief. We defined a utility function based on the mutual information between the state and the information sources. We developed an efficient generalized greedy scheme to iteratively pick observation sources with highest marginal gain, scaled by the added cost. We theoretically established near-optimality of the proposed scheme and further evaluated it on a robotic navigation task.
As part of the future work, we aim to employ PAC greedy maximization~\cite{satsangi2016pac} to accelerate the information selection process since instead of exact computation, it only requires bounds on the utility function.
%%%%%%%%%%%%%%%%%%%%%%%%%%%%%%%%%%%%%%%%%%%%%%%%%%%%%%
%%%%%%%%%%%%%%%%%%%%%%Appendices%%%%%%%%%%%%%%%%%%%%%%
%%%%%%%%%%%%%%%%%%%%%%%%%%%%%%%%%%%%%%%%%%%%%%%%%%%%%%
\normalsize
\begin{appendices}
	%%%%%%%%%%%%%%%%%%%%%%%%%%%%%%%%%%%%%%%%
\section{Proof of Lemma 1}\label{sec:lemma1}

It is clear that $f(\emptyset) = \ent(\srv) - \ent(\srv)=0$. 

Let $[n] = \{1,2,\ldots,n\}$. To prove monotonicity, consider $\iota_1 \subset [n]$ and $j \in [n] \backslash \iota_1$. Then,
\begin{equation*}
	\begin{split}
		\ent(\srv|&\bigcup_{i \in \iota_1\cup\{j\}} \omegarv^i) \\
		\labeq{(a)}& \ent(\bigcup_{i \in \iota_1\cup\{j\}} \omegarv^i|\srv) + \ent(\srv) - \ent(\bigcup_{i \in \iota_1\cup\{j\}} \omegarv^i) \\
		\labeq{(b)}& \ent(\bigcup_{i \in \iota_1} \omegarv^i|\srv) + \ent(\omegarv^j|\srv) + \ent(\srv) - \ent(\bigcup_{i \in \iota_1} \omegarv^i) \\ &- \ent(\omegarv^j|\bigcup_{i \in \iota_1} \omegarv^i) \\
		\labeq{(c)}& \ent(\srv|\bigcup_{i \in \iota_1} \omegarv^i) + \ent(\omegarv^j|\srv) - \ent(\omegarv^j|\bigcup_{i \in \iota_1} \omegarv^i) \\
		\labeq{(d)}& \ent(\srv|\bigcup_{i \in \iota_1} \omegarv^i) + \ent(\omegarv^j|\srv,\bigcup_{i \in \iota_1} \omegarv^i) - \ent(\omegarv^j|\bigcup_{i \in \iota_1} \omegarv^i) \\
		\lableq{(e)}& \ent(\srv|\bigcup_{i \in \iota_1} \omegarv^i) + \ent(\omegarv^j|\bigcup_{i \in \iota_1} \omegarv^i) - \ent(\omegarv^j|\bigcup_{i \in \iota_1} \omegarv^i) \\
		=& \ent(\srv|\bigcup_{i \in \iota_1} \omegarv^i),
	\end{split}
\end{equation*}
where $(a)$ and $(c)$ are due to Bayes' rule for entropy, $(b)$ follows from the conditional independence assumption and joint entropy definition, $(d)$ is due to the conditional independence assumption, and $(e)$ stems from the fact that conditioning does not increase entropy.

Furthermore, from the third line of above proof, we can derive the marginal gain as:
\begin{equation*}
	\begin{split}
		f_{j}(\iota_1) &= \ent(\srv|\bigcup_{i \in \iota_1} \omegarv^i) -  \ent(\srv|\bigcup_{i \in \iota_1\cup\{j\}} \omegarv^i) \\
		&= \ent(\omegarv^j|\bigcup_{i \in \iota_1} \omegarv^i) - \ent(\omegarv^j|\srv)
	\end{split}
\end{equation*}

To prove submodularity, let $\iota_1 \subseteq \iota_2 \subset [n]$ and $j \in [n] \backslash \iota_2$. Then,
\begin{equation*}
	\begin{split}
		f_{j}(\iota_1) &= \ent(\omegarv_j|\bigcup_{i \in \iota_1} \omegarv^i) - \ent(\omegarv_j|\srv) \\
		&\labgeq{(a)} \ent(\omegarv_j|\bigcup_{i \in \iota_1\cup(\iota_2\backslash\iota_1)} \omegarv^i) - \ent(\omegarv_j|\srv) \\
		&\labeq{(b)} \ent(\omegarv_j|\bigcup_{i \in \iota_2} \omegarv^i) - \ent(\omegarv_j|\srv) = f_{j}(\iota_2),
	\end{split}
\end{equation*}
where $(a)$ is based on the fact that conditioning does not increase entropy, and $(b)$ results from $\iota_1 \subseteq \iota_2$.

%%%%%%%%%%%%%%%%%%%%%%%%%%%%%%%%%%%%%%%%
\section{Proof of Theorem 2}\label{sec:thm2}

Let $p^0 := b_b^{' a,\omega}$ to be the updated belief (see \eqref{eq:up1}) after taking action $a$ and receiving observation $\omega$. Also, let $p^g := b_{b'}^{'' a,\iota^g,\bar{\omega}}$ and $p^* := b_{b'}^{'' a,\iota^*,\bar{\omega}}$ to denote the updated beliefs (see \eqref{eq:up2}) after receiving auxiliary observations corresponding to the proposed generalized greedy scheme and the optimal selection, respectively.
First, by leveraging the relation between mutual information and Kullback-Leibler (KL-) divergence, we establish the followings: 
\begin{subequations}\label{eq:in-kl}
	\begin{equation}
		\info(\srv;\bigcup_{i \in \iota^g} \omegarv^i) = 
		\E_{\bigcup_{i \in \iota^g} \omegarv^i}\left[\dkl(p^g \| p^0)\right],
	\end{equation}
	\begin{equation}
		\info(\srv;\bigcup_{i \in \iota^*} \omegarv^i) = 
		\E_{\bigcup_{i \in \iota^*} \omegarv^i}\left[\dkl(p^* \| p^0)\right].
	\end{equation}
\end{subequations}
In other words, the mutual information between the state and a set of information sources is equivalent to expected KL-divergence from current belief to posterior belief. 
Therefore, using \eqref{eq:in-kl} along the result of \thmref{thm:gr-app} yields:
\begin{equation}\label{eq:app-kl}
\begin{aligned}
	\E_{\bigcup_{i \in \iota^g} \omegarv^i}\left[\dkl(p^g \| p^0)\right]
	\ge& \\
	\left(1-\frac{1}{\sqrt{e}}\right)&
	\E_{\bigcup_{i \in \iota^*} \omegarv^i}\left[\dkl(p^* \| p^0)\right].
\end{aligned}
\end{equation}
Next, we use the Pythagorean theorem for KL-divergence~\cite{csiszar1975divergence} and take expectation over all realizations of the observations to obtain:
\begin{equation}\label{eq:pyth}
	\begin{aligned}
	\E_{\bigcup_{i \in \iota^*} \omegarv^i}\left[\dkl(p^*  \| p^0)\right] 
	\ge
	\E_{\bigcup_{i \in [n]} \omegarv^i}\left[\dkl(p^* \| p^g)\right]&
	\\+
	\E_{\bigcup_{i \in \iota^g} \omegarv^i}\left[\dkl(p^g \| p^0)\right]&.
	\end{aligned}
\end{equation}
We combine \eqref{eq:app-kl} and \eqref{eq:pyth}, and rearrange the terms to establish the following:
\begin{equation}\label{eq:app-kl-2}
	\begin{aligned}
		\E_{\bigcup_{i \in [n]} \omegarv^i}\left[\dkl(p^* \| p^g)\right]
		\le
		\frac{1}{\sqrt{e}}
		\E_{\bigcup_{i \in \iota^*} \omegarv^i}\left[\dkl(p^* \| p^0)\right],
	\end{aligned}
\end{equation}
where the right hand side is a constant.
Lastly, we exploit Pinkster's inequality which relates the total variation distance to KL-divergence and apply Jansen's inequality for square-root function (a concave function) to derive the desired result:
\begin{equation*}
	\E [\norm{b^g - b^*}_1] \le 
	\sqrt{\frac{2}{\sqrt{e}}
	\E_{\bigcup_{i \in \iota^*} \omegarv^i}\left[\dkl(p^* \| p^0)\right]}.
\end{equation*}

%%%%%%%%%%%%%%%%%%%%%%%%%%%%%%%%%%%%%%%%
\section{Proof of Theorem 3}\label{sec:thm3}

Let $\alpha^g$ and $\alpha^*$ to represent the gradient of value function at $b^g$ and $b^*$, respectively. Let ${R_{max} = \max_{s,a} R(s,a)}$ and ${R_{min} = \min_{s,a} R(s,a)}$. Therefore, we can show that
\begin{equation*}
	\begin{aligned}
	\E [V(b^g) - V(b^*)] 
	&= \E [\alpha^g.b^g - \alpha^*.b^*] \\
	&= \E [\alpha^g.b^g - \alpha^g.b^* + \alpha^g.b^* -\alpha^*.b^*] \\
	&\lableq{(a)} \E [\alpha^g.b^g - \alpha^g.b^* + \alpha^*.b^* -\alpha^*.b^*] \\
	&= \E [\alpha^g.(b^g - b^*)] \\
	&\lableq{(b)} \E [\norm{\alpha^g}_\infty \norm{b^g - b^*}_1] \\
	&\lableq{(c)} \delta \frac{\max\left\{|R_{max}|,|R_{min}|\right\}}{1-\gamma},
	\end{aligned}
\end{equation*}
where $(a)$ follows from the fact that $\alpha^*$ is the gradient of optimal value function, $(b)$ is due to H\"{o}lder's inequality, and $(c)$ is the result of \thmref{thm:dist} and the fact that ${\norm{\alpha}_\infty \le \frac{\max\left\{|R_{max}|,|R_{min}|\right\}}{1-\gamma}}$ for every $\alpha$ vector.
\end{appendices}
%%%%%%%%%%%%%%%%%%%%%%%%%%%%%%%%%%%%%%%%%%%%%%%%%%%%%%%%
%%%%%%%%%%%%%%%%%%%%%%%References%%%%%%%%%%%%%%%%%%%%%%%
%%%%%%%%%%%%%%%%%%%%%%%%%%%%%%%%%%%%%%%%%%%%%%%%%%%%%%%%
\bibliographystyle{ieeetr}\footnotesize
\bibliography{main}

\end{document}